\documentclass{article}

    \PassOptionsToPackage{numbers, compress}{natbib}

    \usepackage[final]{neurips_2024}

\usepackage[utf8]{inputenc} 
\usepackage[T1]{fontenc}    
\usepackage{hyperref}       
\usepackage{url}            
\usepackage{booktabs}       
\usepackage{amsfonts}       
\usepackage{nicefrac}       
\usepackage{microtype}      
\usepackage{xcolor}         
\usepackage{graphicx}       
\usepackage{amsmath}
\title{Quantifying Positional Biases in Text Embedding Models}

\author{%
  Reagan J. Lee$^*$ \\
  University of California, Berkeley \\
  Berkeley, CA 94704 \\
  \texttt{reaganjlee@berkeley.edu} \\
  \And
  Samarth Goel$^*$ \\
  University of California, Berkeley\\
  Berkeley, CA 94704 \\
  \texttt{sgoel9@berkeley.edu} \\
  \AND
  Kannan Ramchandran \\
  University of California, Berkeley \\
  Berkeley, CA 94704  \\
  \texttt{kannanr@eecs.berkeley.edu} \\
}

\begin{document}

\maketitle

\def\thefootnote{*}\footnotetext{Equal contribution.}

\begin{abstract}

Embedding models are crucial for tasks in Information Retrieval (IR) and semantic similarity measurement, yet their handling of longer texts and associated positional biases remains underexplored. In this study, we investigate the impact of content position and input size on text embeddings. Our experiments reveal that embedding models, irrespective of their positional encoding mechanisms, disproportionately prioritize the beginning of an input. Ablation studies demonstrate that insertion of irrelevant text or removal at the start of a document reduces cosine similarity between altered and original embeddings by up to 12.3\% more than ablations at the end. Regression analysis further confirms this bias, with sentence importance declining as position moves further from the start, even with with content-agnosticity. We hypothesize that this effect arises from pre-processing strategies and chosen positional encoding techniques. These findings quantify the sensitivity of retrieval systems and suggest a new lens towards embedding model robustness.\footnote{Code available: \url{https://github.com/sgoel97/neurips-embedding-positional-bias}}

\end{abstract}

\section{Introduction}

Embedding models are increasingly used to encode text in critical applications like document search systems. However, their effectiveness diminishes when dealing with long-context inputs, particularly in larger documents that cannot entirely fit into these models' context windows. To address these limitations, techniques such as document chunking are used to segment large documents into smaller pieces of text as model inputs \cite{zhu2024longembed}. Despite its utility, research into optimal chunking strategies is still an emerging field and improvements can often be highly domain-specific or underexplored in practical environments. \cite{zhang2024raft}.

In this study, we investigate the influence of content position and input size on the resulting text embedding vector from eight embedding models. Our findings reveal a systematic bias in which embedding models, regardless of their positional encoding mechanisms, disproportionately weigh the beginning of a text input. This results in greater importance being assigned to the initial sentences of multi-sentence or long-context inputs. To demonstrate this, we conducted two types of ablation studies: one involving the insertion of irrelevant text ("needles") at different positions in the document \cite{guerreiro2023looking}, and another involving the removal of varying text chunks. We observe that inserting irrelevant text at the beginning of a document reduces the cosine similarity between the altered and original document embeddings by up to 8.5\% more than when inserted in the middle, and 12.3\% more than when inserted at the end. Similarly, removal experiments show that the largest decreases in similarity occur when text is removed from the beginning of the document.

To further explore this bias, we employ regression analysis to measure sentence-level importance on a complete document-level embedding, isolating model position bias from human writing patterns. Our analysis shows a significant decline in regression coefficients as the sentence position moves further from the beginning of the document, reinforcing the bias toward earlier content. To rule out dataset-specific effects, we repeat all experiments with randomly shuffled sentences and obtain similar results, confirming that this bias arises from the model's internal mechanisms rather than document structure.

We hypothesize that this bias stems from common pre-processing strategies, particularly truncation, used during training when the input exceeds the model's context window \cite{liu2019roberta, bge_embedding}. This has important implications for real-world retrieval tasks, where documents with key information located later in the text may be overlooked due to the model's disproportionate weighting of early content \cite{barnett2024seven}. 

We conclude by discussing the broader implications of these biases in embedding models and highlight the need for future research to develop methods that can better handle the entirety of long-context inputs without disproportionately prioritizing the beginning.
 
\section{Background}

\subsection{Bidirectional encoding in embedding models}

Embedding models, particularly those utilizing transformer encoder architectures \cite{vaswani2023attention}, employ layers of bidirectional self-attention blocks to process text \cite{devlin2019bert}. These models are distinct from decoders in that they generate a fixed-length vector representing the entire input text. This is achieved by producing an output matrix \( L \times D \) (where \( L \) is the sequence length and \( D \) is the dimensionality of the embeddings), and then applying either mean or max pooling across the \( L \) dimension \cite{reimers2019sentencebert}. Such pooling operations are position-invariant, theoretically suggesting an unbiased treatment of input positions in terms of attention and representation \cite{su2023roformer}.

We use cosine similarity to compare the output embeddings from these models, especially to study the effects of textual modifications such as insertions or deletions. Cosine similarity measures the cosine of the angle between two vectors, thus providing a scale- and orientation-invariant metric to assess the similarity between two text representations \cite{li2024angleoptimized}. Due to the invariance of the architecture and similarity measurement we employ, the last systematic source of bias stems from learned positional embeddings used in our models and the models' training methodology, which are heavily connected.

\subsection{Positional Encoding Techniques}

\textbf{Absolute Positional Embedding (APE)} assigns fixed position-specific vectors based off of position id to each token embedding. This was first popularized by BERT \cite{devlin2019bert} and remains the most common technique to add positional information in encoder-style models today.

\textbf{Rotary Positional Embedding (RoPE)}: RoPE encodes positions by applying a rotation to each token's embedding in the 2D subspaces of the embedding space. For each embedding vector $x$, it applies a rotation matrix $R(\theta)$ based on the position $pos$:

\[
\mathbf{x}_{\text{pos}}^{(2i)} = \mathbf{x}^{(2i)} \cos(\theta_{\text{pos}}) - \mathbf{x}^{(2i+1)} \sin(\theta_{\text{pos}})
\]
\[
\mathbf{x}_{\text{pos}}^{(2i+1)} = \mathbf{x}^{(2i)} \sin(\theta_{\text{pos}}) + \mathbf{x}^{(2i+1)} \cos(\theta_{\text{pos}})
\]
where \(\theta_{\text{pos}} = \frac{\text{pos}}{10000^{2i/d}}\), \(i\) indexes the embedding dimensions, and \(d\) is the dimensionality.

3. \textbf{Attention with Linear Biases (ALiBi)}: ALiBi introduces a relative bias into the attention scores rather than modifying the embeddings. The bias is linear with respect to the distance between tokens. The attention score \(A(i, j)\) between token \(i\) and token \(j\) is modified by adding a bias term \(m(|i - j|)\), where \(|i - j|\) is the distance between tokens:

\[
A(i, j) = \frac{\mathbf{q}_i \cdot \mathbf{k}_j}{\sqrt{d_k}} + m(|i - j|)
\]

where \(m(|i - j|)\) is a linear function of the relative distance between tokens \(i\) and \(j\), and \(d_k\) is the dimensionality of the key vectors.

\subsection{Noise from Document Chunking for IR Tasks}

In practical applications, documents often exceed the context length capabilities of embedding models, necessitating chunking strategies like naive, recursive, or semantic chunking \cite{fei2023extending, gao2024retrievalaugmented}. This process divides a document into smaller pieces that fit within a model's context window, then embeds each chunk separately for insertion into a vector database \cite{johnson2017billionscale} and downstream use in Retrieval-Augmented Generation (RAG) \cite{lewis2021retrievalaugmented} tasks. This causes an unintentional, outsized amount of noise in the beginning and end of documents as a function of selected chunking strategies.

\subsection{Embedding Models Robustness}

The performance of decoder models has been shown to vary significantly with the position of content within the model's context window, with pronounced degradation observed for inputs that exceed the context length seen during training \cite{liu2023lost}. Positional encoding methods have been studied to address these challenges from both decreasing the effect of content position within training context length\cite{Zhang2024found}, and generalizing to longer contexts from itself\cite{Kazemnejad2023positional}. However, these works exhibit limitations: The former provides limited analysis of diverse encoding mechanisms, and the latter emphasizes generalization to longer inputs rather than robustness to positional shifts.

Moreover, both studies focus exclusively on decoder-only architectures, whose causal attention mask provides the ability for the model to generalize without explicit positional information itself\cite{Kazemnejad2023positional}, and remains underexplored as a research direction. Existing work on embedding model robustness predominantly centers on improving training data quality or diversity\cite{Yang2022robust}, with relatively little attention paid to architectural components such as positional encoding mechanisms.

\section{Effect of sentence-level positioning in embedding output}

We explore how the position and size of a sentence in a text influence a document's final embedding vector. Our methodology adapts the needle-in-a-haystack test \cite{guerreiro2023looking}, traditionally used for generative models in information retrieval \cite{geminiteam2024gemini}, to evaluate embedding models.

\subsection{Experimental setup}

We investigate the impact of adding irrelevant or adversarial text ("needle") to a document. After inserting the needle, we generate a new embedding for the altered text and compare it to the original using cosine similarity. We vary the needle's length (5\%, 10\%, 25\%, 50\%, and 100\% of the original text's token count) and position (beginning, middle, end) across 15 experimental conditions. We use an extended version of Lorem Ipsum placeholder text \cite{timmer2022tsm} that exceeds the length of our longest datapoint and is structured in paragraph format to achieve a needle with structural similarity to our data while avoiding a confounding effect on the embedding model.

In a parallel experiment, we remove portions of text (10\%, 25\%, 50\% of sentences, rounded up) from different positions (beginning, middle, end) in the document. The resulting text is then embedded, and its similarity to the original embedding is measured using cosine similarity.
We test various models, segmented by their positional encodings, to demonstrate the consistency of our results across multiple popular embedding models. We used six open-source models utilizing various positional encoding methods - BGE-m3 \cite{chen2024bge} and E5-Large-V2 \cite{wang2022weak} using APE; Nomic-Embed-Text-v1.5 \cite{nussbaum2024nomic} and E5-RoPE base \cite{zhu2024longembed} using RoPE; and Jina-Embeddings-v2-Base \cite{günther2024jina} and Mosaic-Bert-Base (sequence length 1024) \cite{press2022train} using ALiBi. We additionally test Cohere's Embed-English-v3.0 \cite{reimers2023} due to their popularity and real-world applicability. Although we picked these models due to their varying positional encoding methods and performance, we acknowledge these may not generalize to other architectures and datasets. Context lengths or additional information such as parameter counts or benchmark performance for these models can be found in Appendix \ref{modeldetails}. For texts exceeding these limits, we truncate from the end to fit the models' context windows. For datasets, we use 200 examples each from the PubMed Publications \cite{cohan2018discourseaware}, Paul Graham Essay Collection \cite{samarth_goel_2024}, Amazon Reviews \cite{zhang2016characterlevel}, Argumentative Analysis \cite{wachsmuth-etal-2018-retrieval}, and Reddit Posts \cite{geigle2021tweac} datasets, selected for their range of writing categorizations and lengths. More details on these datasets can be found in appendix \ref{datasetdetails}.

\subsection{Results and discussion}

\begin{figure}[h]
    \centering
    \includegraphics[width=0.7\textwidth]{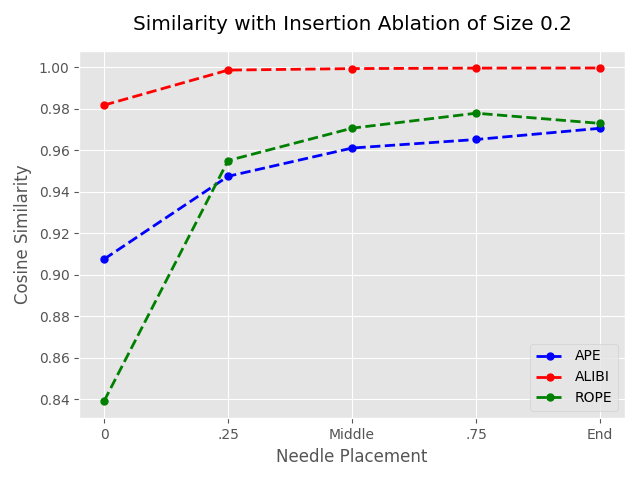}
    \caption{Cosine similarity vs. needle size and position}
    \label{fig:insertion}
\end{figure}
Our results indicate a pronounced drop in similarity when irrelevant text is inserted at the beginning of documents, with less impact observed when additions occur in the middle or end. Specifically, for APE models, introducing an insertion equal to 20\% of the total content at the beginning results in an average cosine similarity of 0.885, compared to 0.963 at the end—a relative decrease of approximately 8\%. RoPE-based models show a stronger sensitivity to this disruption, with cosine similarity dropping to 0.819 at the beginning, a 15.4\% decrease compared to the 0.968 similarity at the end. By contrast, AliBi models are the most robust, maintaining a high cosine similarity of 0.981 at the beginning and 0.999 at the end, reflecting only a 1.8\% decrease. This suggests that earlier positions in the input sequence play a more critical role in model performance, and different positional encoding methods vary in their resilience to this type of input perturbation.

This trend persists across all insertion sizes, with larger insertions intensifying the drop in similarity. Even though the magnitude of the degradation varies by model, we find the trend robust to model differences. Across all five models tested, the average decrease in cosine similarity is approximately 7\%, indicating a consistent pattern of sensitivity to input alterations at the beginning of the sequence.

Additionally, we observe that removal ablations yield similar results, although the overall similarity scores are higher in comparison to insertion ablations. This suggests that while the models are affected by both insertion and removal disruptions, the impact of irrelevant insertions at the beginning of sequences may introduce greater noise into the representations.

Similar trends are observed in the removal experiments, where the largest impacts on similarity occur when sentences are removed from the beginning. Removing half of the sentences from the beginning results in a median similarity that is 10.6\% lower than when sentences are removed from the end, with no significant difference between middle and end removals—unlike the insertion experiments. Interestingly, even a 50\% text removal from the middle maintains a median similarity of 95\%, corroborating our findings from the insertion experiments, where a large drop in similarity was expected but not observed. These results suggest that while the position of removed content has a clear impact, it is somewhat less disruptive than insertions. 

\section{Analysis of embedding decomposition}

Recent advancements in embedding interpretability have demonstrated that certain dimensions in high-dimensional semantic spaces may correspond to specific linguistic or semantic features, such as sentiment or subject matter \cite{dar2023analyzing}. Further research has shown that vector operations, such as adding embeddings, can produce new vectors that represent the semantic meaning of their components \cite{Senel_2018}.

Building from these works, we explore the impact of sentence-level positioning on the final document embedding vector through regression analysis, which offers a more direct method to quantify the contribution of individual sentences to a document's embedding representation. 

Human writing often emphasizes key information at the beginning and end of documents, a technique that may introduce biases in datasets and reason for embeddings to skew towards these positions. To address these, we employ additional data augmentation and ablation techniques aimed at isolating and understanding these effects, to ensure that our findings more accurately reflect model behavior rather than dataset peculiarities.

\subsection{Reconstructing embedding vectors through linear combinations of constituents}

To start, we wanted to validate the assumption that the sentence embeddings of a larger document can meaningfully be used as a proxy for the original document embedding \cite{tsukagoshi2022comparison}. To test this, we wanted to determine how much reconstruction loss we would incur from using an optimal linear combination of sentence embedding vectors instead of a full multi-sentence embedding vector. Optimizing for train $R^2$, we use Ordinary Least Squares (OLS) regression to reconstruct the document embedding from its sentence embeddings, with the multi-sentence embedding vector as our response and each sentence vector as a predictive datapoint for our regression. Our model choice is notable for its direct interpretability \cite{słoczyński2020interpreting}, though we acknowledge and check for potential issues posed by OLS, such as multicollinearity. Our regressions use normalized embeddings (L2 norm of 1) to ensure scale invariance \cite{Steck_2024}. We separate our data points into their component sentences by use of punctuation such as periods, and new lines.

When we regress the sentence embedding vectors onto the multi-sentence embedding vector, we find that our train $R^2$ across the eight models and five datasets we used ranges from 0.75 to 0.99, with an average $R^2$ or 0.876 when reconstructing the multi-sentence embedding vector. This result indicates that approximately 87.6\% of the variance in a long-content document embedding can be accounted for by analyzing the embeddings of the individual sentences constituting the document. The Mean Squared Error (MAE) summed over all dimensions of this reconstruction across all models and datasets ranged from 0.001 and 0.01 with an average of 0.0069, suggesting minimal deviation in the reconstructed vectors.

\subsection{Analyzing regression coefficients as importance weights}

Given the high explanatory power of our regression models, the coefficients given to each sentence (datapoint) in our regression are strong indicators to determine their relative importance to the total document. To standardize our comparisons across documents, we standardized each coefficient vector by its L2 norm. One potential issue to note with this approach is the presence of negative coefficient values, but these tended to be rare and very low in magnitude, with very little influence on our final analysis.

We judge the importance of a sentence by its regression coefficient. For example, if a regression on a two-sentence document yielded weights 0.8 and 0.6, we conclude that the first sentence is 33.3\% more important to the final semantic meaning of the text than the second sentence.

 \begin{figure}
    \centering
    \includegraphics[width=\textwidth]{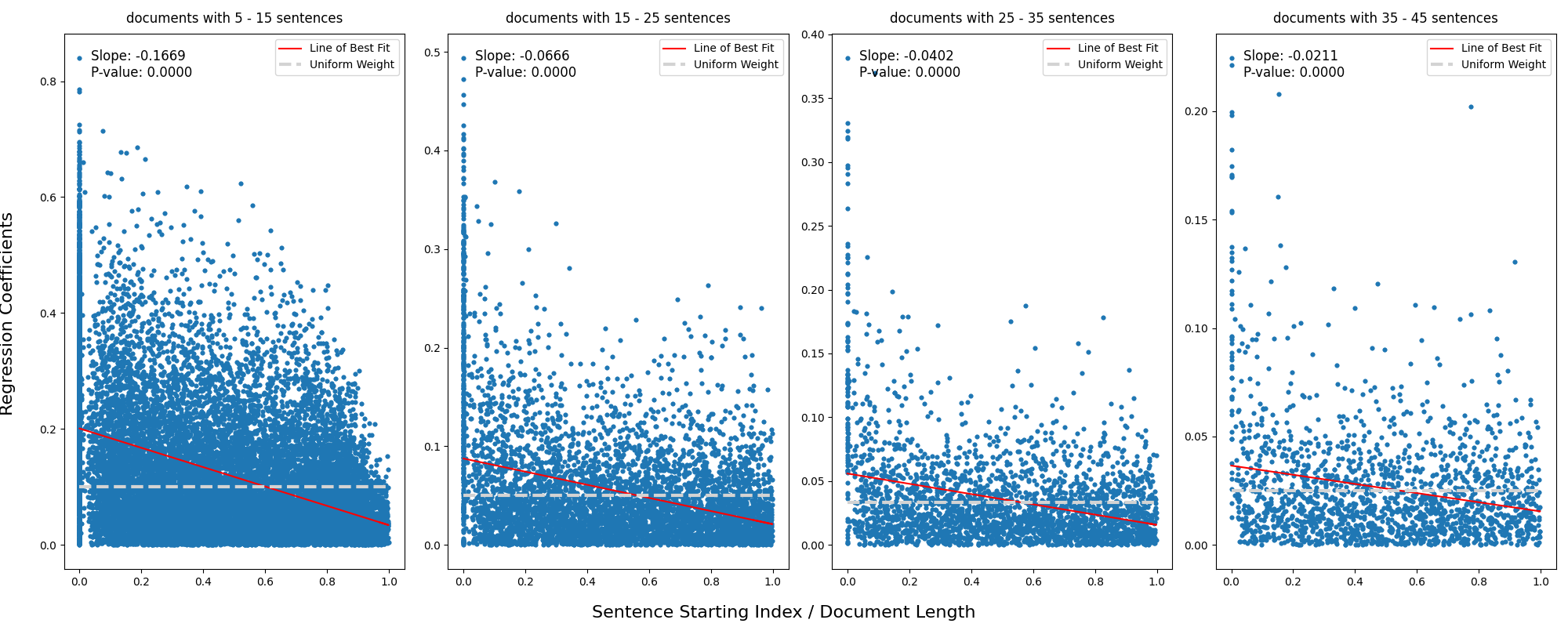}
    \caption{Regression coefficients vs. sentence position, bucketed by document length}
    \label{fig:r2}
\end{figure}

As shown in Figure 2, there is a downward trend in coefficient values with increasing sentence position, suggesting a positional bias where earlier sentences generally have a greater impact on the document's overall semantic representation. To quantify this observation, we plot regression coefficients against sentence positions over all the documents in our dataset. 

\begin{table}[ht]
\centering
\caption{Correlation and statistical significance of sentence position against shuffled text}
    \begin{tabular}{@{}ccc}
    \toprule
    Positional Encoding & \small\textbf{Correlation}& \small\textbf{P-value}\\
    \midrule
APE& -0.127657& 2.233374e-103\\
RoPE& -0.115861& 2.259581e-85\\
ALiBi& -0.07615& 9.205763e-38\\
    \end{tabular}
    \label{table1}
\end{table}

\subsection{Embedding positional bias is robust to human-level writing bias}

To validate that this observed bias is not solely a byproduct of dataset-specific characteristics, namely human-level writing bias, we conducted additional regression experiments where all sentences from the above pre-processing steps were shuffled before their embeddings were generated. Using these new embeddings, remarkably, the results mirrored the original findings, with the randomly selected first sentence in the shuffled document consistently receiving a higher weight, thereby disambiguating our results from potential dataset biases.

More specifically, we expect the weight assigned to the first sentence to follow a uniform weight of \( \frac{1}{\text{num\_sentences}} \). However, this analysis shows a distinct negative correlation between sentence position and importance score, with significant deviations from the expected uniform distribution (\( \alpha \ll 0.001 \)), confirming a systematic positional influence within document embeddings as shown in table \ref{table1}. These findings suggest that the embedding models may inherently prioritize the initial information presented in any text sequence, irrespective of its original position in the document.

\section{Isolating the role of training methodology in model biases}

During training, input data is processed sequentially, starting at the beginning of the context window. Variable-length training samples are packed into this fixed window, often necessitating truncation when the input exceeds the window’s length. Truncation typically discards content from the end, leading to a systematic bias where earlier positions in the sample receive disproportionate attention.

For a given position \(i \in \) [0, N] within a context window of length \(N\), the model observes \(t_i \), the number of non-padding tokens encountered at position \(i \). The importance of position \(i \) can then be modeled as  \(imp(t_i)\) = \(u(t_i)\), where \(u(\cdot)\) represents the model's updates based on the presence of non-padding tokens at \(t_i\).

As traditional truncation favors earlier positions, the frequency with which tokens are seen at the beginning of the context window is inherently higher than at the end. This can be modeled as a monotonically decreasing function, where the quantity of non-padding tokens at \(t_i\) diminishes as \(i\) increases. As a result, the relative importance of earlier positions \(imp(t_1) \geq imp(t_2) \geq  \dots \geq imp(t_N)\) is systematically higher, introducing an implicit bias that prioritizes early context over later content.

Although this monotonic impact on position can theoretically be removed by maintaining an equal number of effective updates throughout the context, it is unknown what the impacts on computational costs, and model performance would be. Future pre-training, as well as employing novel context-length enhancement methods, with this bias in mind will require additional research to fully understand the impacts, leading us to believe that this bias will continue in future models.

\section{Conclusion}

Our study uncovers a positional bias in embedding models, where sentences at the beginning of a document disproportionately influence the resulting embeddings. This bias is consistently observed across various models with different context sizes and datasets and is evident in both text insertion and removal experiments. We further quantified this effect through regression analysis, which highlights the extent of the model's preference for earlier content. Our findings suggest that this bias is intrinsic to the models’ training methodologies, particularly the use of truncation strategies, rather than a consequence of dataset-specific patterns.

This positional bias poses challenges in critical applications like information retrieval in document search systems, highlighting the need for alternative positional encoding methods to mitigate these biases and achieve more balanced semantic representations. Additionally, growing research into extending context lengths offers a promising avenue for further exploration of this phenomenon and potential solutions.

\bibliography{References}








\appendix

\section{Model details}\label{modeldetails}

\paragraph{Embed-English-v3.0 \cite{reimers2023}} has a content length of 512 tokens and an unknown number of parameters. The model is accessed via the Cohere API.

\paragraph{BGE-m3 \cite{chen2024bge}} has a content length of 8912 tokens, is comprised of 568M parameters, and was trained using the APE positional encoding method.

\paragraph{E5-Large-V2 \cite{wang2022weak}} has a content length of 512 tokens, is comprised of 335M parameters, and was trained using the APE positional encoding method.

\paragraph{Nomic-Embed-Text-v1.5 \cite{nussbaum2024nomic}}  has a content length of 8192 tokens, is comprised of 137M parameters, and was trained using the RoPE positional encoding method.

\paragraph{E5-RoPE-base \cite{zhu2024longembed}}  has a content length of 512 tokens, is comprised of 108M parameters, and was trained using the RoPE positional encoding method.

\paragraph{Jina-Embeddings-v2-Base\cite{günther2024jina}}  has a content length of 8192 tokens, is comprised of 137M parameters, and was trained using the ALiBi positional encoding method.

\paragraph{Mosaic-Bert-Base \cite{press2022train}}  has a content length of 1024 tokens, is comprised of 110M parameters, and was trained using the ALiBi positional encoding method.

\section{Dataset details}\label{datasetdetails}

\paragraph{PubMed Publications \cite{cohan2018discourseaware}:} We use PubMed publication abstracts to assess the impact of our ablations on scientific writing. Scientific texts are characterized by their structured presentation of information and specialized vocabulary. Understanding how embeddings capture this complexity can provide insights into their utility in academic and research applications. This dataset is comprised of 270,000 datapoints.

\paragraph{Paul Graham Essay Collection \cite{samarth_goel_2024}:} We analyze over 200 essays written by Paul Graham, varying from 400 to 70,000 words. Paul Graham’s essays are known for their thoughtful, reflective style and coherent argument structure, making them ideal for studying how embeddings handle nuanced and complex idea development over long texts. This dataset is comprised of 215 datapoints.

\paragraph{Amazon Reviews \cite{zhang2016characterlevel}:} Drawn from MTEB’s Amazon Polarity dataset, this helps us examine consumer review text. Reviews are direct and opinion-rich, offering a perspective on how embeddings process everyday language and sentiment, which is crucial for applications in consumer analytics. This dataset is comprised of 4 million datapoints.

\paragraph{Argumentative Analysis \cite{wachsmuth-etal-2018-retrieval}:} From the BiER benchmark’s Argumentative Analysis (ArguAna) dataset, we explore embeddings of formal persuasive writing. This dataset includes well constructed arguments that are ideal for testing how embeddings capture logical structure and the effectiveness of rhetoric. This dataset is comprised of 10,000 datapoints.

\paragraph{Reddit Posts \cite{geigle2021tweac}:} More Informal and diverse writing styles can be found on Reddit. This dataset introduces grammar, style, and subject matter diversity into our tests, extending our findings to be more robust and adaptable to a wide range of writing styles. This dataset is comprised of 450,000 datapoints.

\section{Cosine similarities across insertion ablation sizes and datasets}

The following are the results of running insertion and removal ablations of given sizes on input examples. These are the results of the average cosine similarity across all datasets. 

\begin{figure}[h]
    \centering
    \begin{minipage}{0.47\textwidth}
        \centering
        \includegraphics[width=\textwidth]{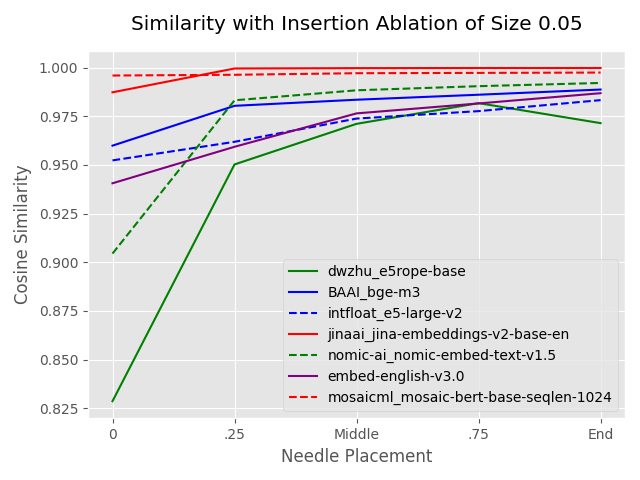}
        \label{fig:cosine_0.05_insert}
    \end{minipage}\hfill
    \begin{minipage}{0.47\textwidth}
        \centering
        \includegraphics[width=\textwidth]{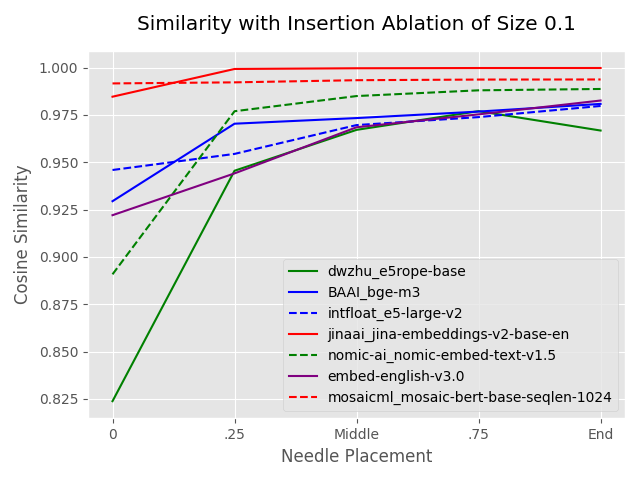}
        \label{fig:levenshtein}
    \end{minipage}
\end{figure}

\begin{figure}[h]
    \centering
    \begin{minipage}{0.47\textwidth}
        \centering
        \includegraphics[width=\textwidth]{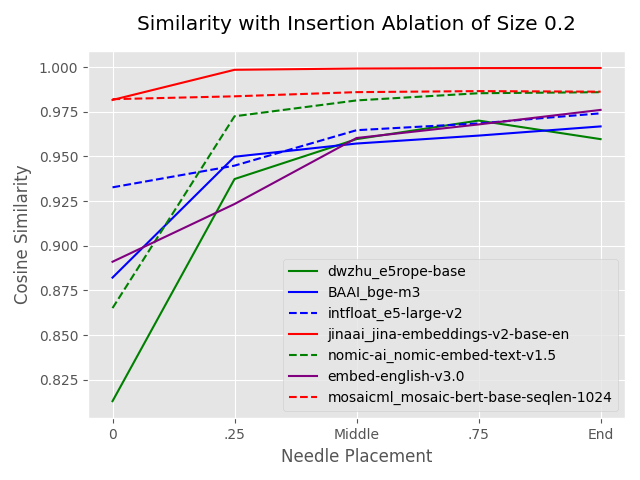}
        \label{fig:cosine_0.2_insert}
    \end{minipage}\hfill
    \begin{minipage}{0.47\textwidth}
        \centering
        \includegraphics[width=\textwidth]{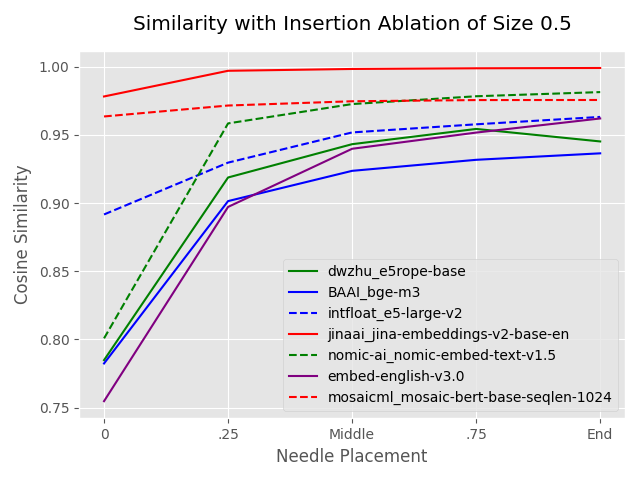}
        \label{fig:levenshtein}
    \end{minipage}
\end{figure}

\begin{figure}[h]
    \centering
    \begin{minipage}{0.47\textwidth}
        \centering
        \includegraphics[width=\textwidth]{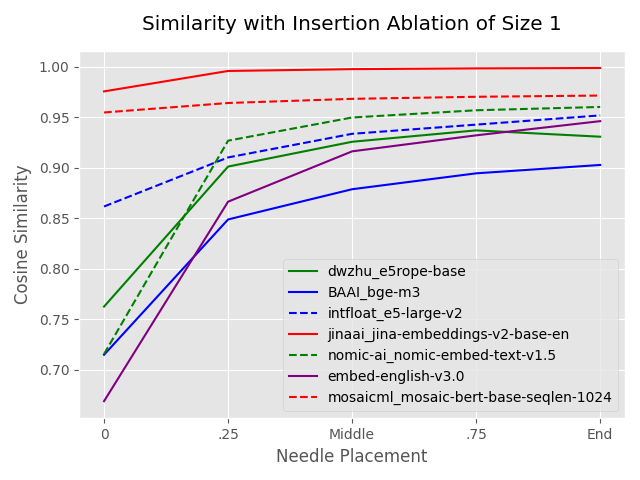}
        \label{fig:cosine}
    \end{minipage}
\end{figure}

\section{Cosine similarities across deletion of ablation sizes and datasets}

\begin{figure}[h]
    \centering
    \begin{minipage}{0.47\textwidth}
        \centering
        \includegraphics[width=\textwidth]{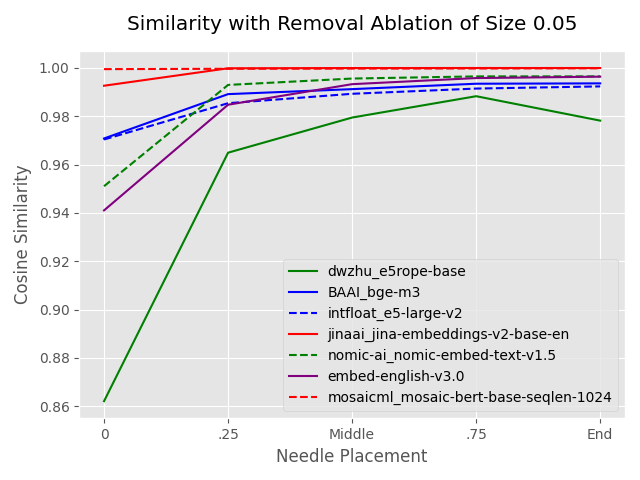}
        \label{fig:cosine_0.05_remove}
    \end{minipage}\hfill
    \begin{minipage}{0.47\textwidth}
        \centering
        \includegraphics[width=\textwidth]{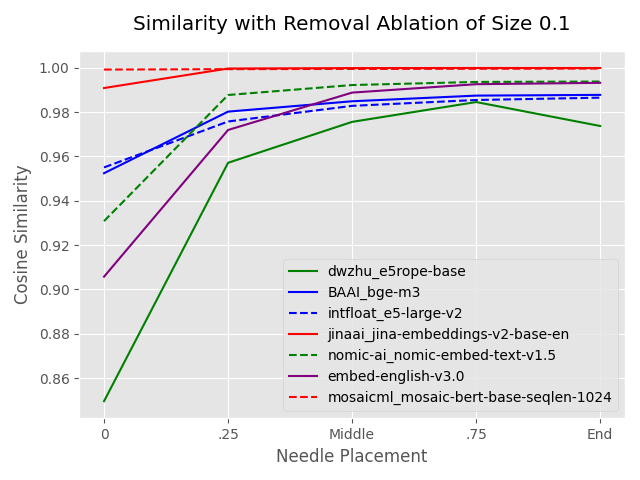}
        \label{fig:levenshtein}
    \end{minipage}
\end{figure}

\begin{figure}[h]
    \centering
    \begin{minipage}{0.47\textwidth}
        \centering
        \includegraphics[width=\textwidth]{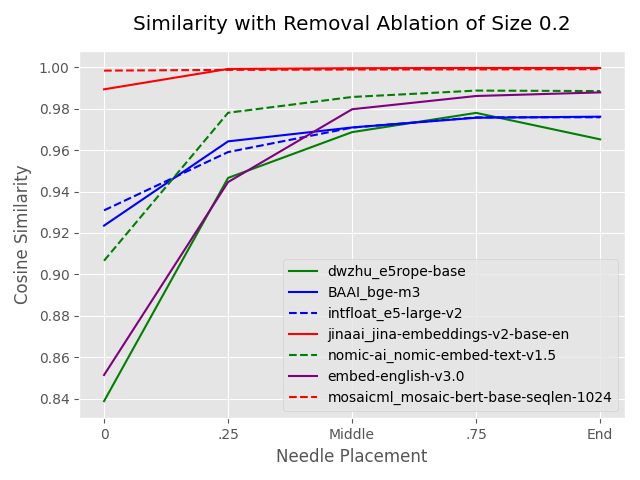}
        \label{fig:cosine_0.2_remove}
    \end{minipage}\hfill
    \begin{minipage}{0.47\textwidth}
        \centering
        \includegraphics[width=\textwidth]{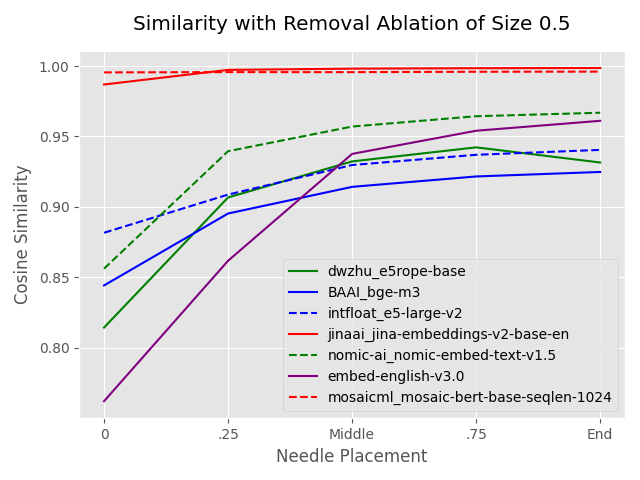}
        \label{fig:levenshtein}
    \end{minipage}
\end{figure}
\section{Sentence Position Against Shuffled Text}

Three sentence length range buckets (65-75, 75-85, 95-105) were omitted due to small sample size (n=6). Examples with less than 5 sentences each were omitted. 
\begin{table}[ht]
\centering
\caption{ALiBi}
    \begin{tabular}{@{}cccc}
    \toprule
    Sentence Length Range& \small\textbf{Correlation}& \small\textbf{P-value} & \small\textbf{Number of Samples}\\
    \midrule
5-15& -0.120560& 1.037594e-24& 904\\
15-25& -0.083780& 2.757708e-05& 132\\
25-35& -0.015695& 5.596307e-01& 48\\
 35-45& -0.037581& 5.387906e-02& 66\\
 45-55& -0.008077& 4.455038e-01& 178\\
 55-65& -0.019355& 1.426657e-01& 98\\
    \end{tabular}
    \label{shuffle-alibi}
\end{table}

\begin{table}[ht]
\centering
\caption{APE}
    \begin{tabular}{@{}cccc}
    \toprule
    Sentence Length Range& \small\textbf{Correlation}& \small\textbf{P-value} & \small\textbf{Number of Samples} \\
    \midrule
5-15& -0.204936& 1.196681e-88& 904\\
15-25& -0.123513& 1.420863e-18& 132 \\
25-35& -0.036560& 2.585037e-02& 48 \\
 35-45& -0.034370& 7.942209e-04& 66 \\
 45-55& -0.009526& 5.458451e-02& 178 \\
 55-65& -0.004620& 4.229611e-01& 98 \\
    \end{tabular}
    \label{shuffle-ape}
\end{table}

\begin{table}[ht]
\centering
\caption{RoPE}
    \begin{tabular}{@{}cccc}
    \toprule
    Sentence Length Range& \small\textbf{Correlation}& \small\textbf{P-value} & \small\textbf{Number of Samples} \\
    \midrule
5-15& -0.201598& 3.154022e-69& 904\\
15-25& -0.098903& 1.669302e-11& 132 \\
25-35& -0.044463& 8.444829e-03& 48 \\
 35-45& -0.021359& 4.203218e-02& 66 \\
 45-55& -0.009357& 6.387572e-02& 178 \\
 55-65& -0.008881& 1.290475e-01& 98 \\
    \end{tabular}
    \label{shuffle-rope}
\end{table}

\end{document}